\documentclass[10pt,twocolumn,letterpaper]{article}

\usepackage[pagenumbers]{cvpr} % To force page numbers, e.g. for an arXiv version

\usepackage{capt-of}
\usepackage[dvipsnames]{xcolor}
\usepackage{xspace}
\usepackage{enumitem}
\usepackage{amsmath,amssymb,amsbsy,amsfonts,dsfont,pifont,bm,bbm,mathrsfs,mathtools,nicefrac}
\usepackage{algorithm,algpseudocode,listings}
\usepackage{booktabs,multirow,adjustbox,diagbox,threeparttable,tabularray,colortbl}

\usepackage{indentfirst} % added by wx

\newcommand{\method}{\texttt{TF-T2V}\xspace}

\newcommand{\tocite}[1]{{\color{red} [TO CITE]}}

\definecolor{Gray}{gray}{0.94}
\definecolor{liGray}{gray}{0.5}
\definecolor{LightCyan}{rgb}{0.88,1,1}

\usepackage{colortbl} 
\usepackage{xcolor}

\definecolor{cvprblue}{rgb}{0.21,0.49,0.74}
\usepackage[pagebackref,breaklinks,colorlinks,citecolor=cvprblue]{hyperref}
\usepackage{wrapfig}
\usepackage[capitalize]{cleveref}  % Should be loaded after 'hyperref', and works perfectly with 'subfigure'.
\crefname{section}{Sec.}{Secs.}
\Crefname{section}{Section}{Sections}
\crefname{table}{Tab.}{Tabs.}
\Crefname{table}{Table}{Tables}
\crefname{figure}{Fig.}{Figs.}
\Crefname{figure}{Figure}{Figures}
\crefname{equation}{Eq.}{Eqs.}
\Crefname{equation}{Equation}{Equations}
\newcommand{\tablestyle}[2]{\setlength{\tabcolsep}{#1}\renewcommand{\arraystretch}{#2}\centering\small}

\newlength\savewidth\newcommand\shline{\noalign{\global\savewidth\arrayrulewidth
  \global\arrayrulewidth 1pt}\hline\noalign{\global\arrayrulewidth\savewidth}}

\def\paperID{1315}
\def\confName{CVPR}
\def\confYear{2024}

\title{A Recipe for Scaling up Text-to-Video Generation with Text-free Videos}

\author{
   Xiang Wang$^{1*}$
     \hspace{0.02cm} 
    Shiwei Zhang$^{2}$
     \hspace{0.02cm} 
    Hangjie Yuan$^{3}$
     \hspace{0.02cm} 
    Zhiwu Qing$^1$ 
     \hspace{0.02cm}
    Biao Gong$^2$ 
     \hspace{0.02cm}
    Yingya Zhang$^2$ 
     \hspace{0.02cm} \\
    Yujun Shen$^4$ 
     \hspace{0.02cm}
    Changxin Gao$^1$ 
     \hspace{0.02cm}
     Nong Sang$^{1}$
      \vspace{2mm}
     \\
    $^1$Key Laboratory of Image Processing and Intelligent Control,\\   School of Artificial Intelligence and Automation, Huazhong University of Science and Technology\\
      $^2$Alibaba Group \hspace{0.5cm} $^3$Zhejiang University \hspace{0.5cm} $^4$Ant Group 
      \vspace{-1mm}
      \\
    \vspace{-2mm}
{\tt\scriptsize \{wxiang,qzw,cgao,nsang\}@hust.edu.cn, \{zhangjin.zsw,yingya.zyy\}@alibaba-inc.com}\\ 
{\tt\scriptsize hj.yuan@zju.edu.cn, \{a.biao.gong,shenyujun0302\}@gmail.com}
}

\begin{document}

\twocolumn[{
\maketitle
    \centering
    \vspace{-25pt}
    \includegraphics[width=0.95\textwidth]{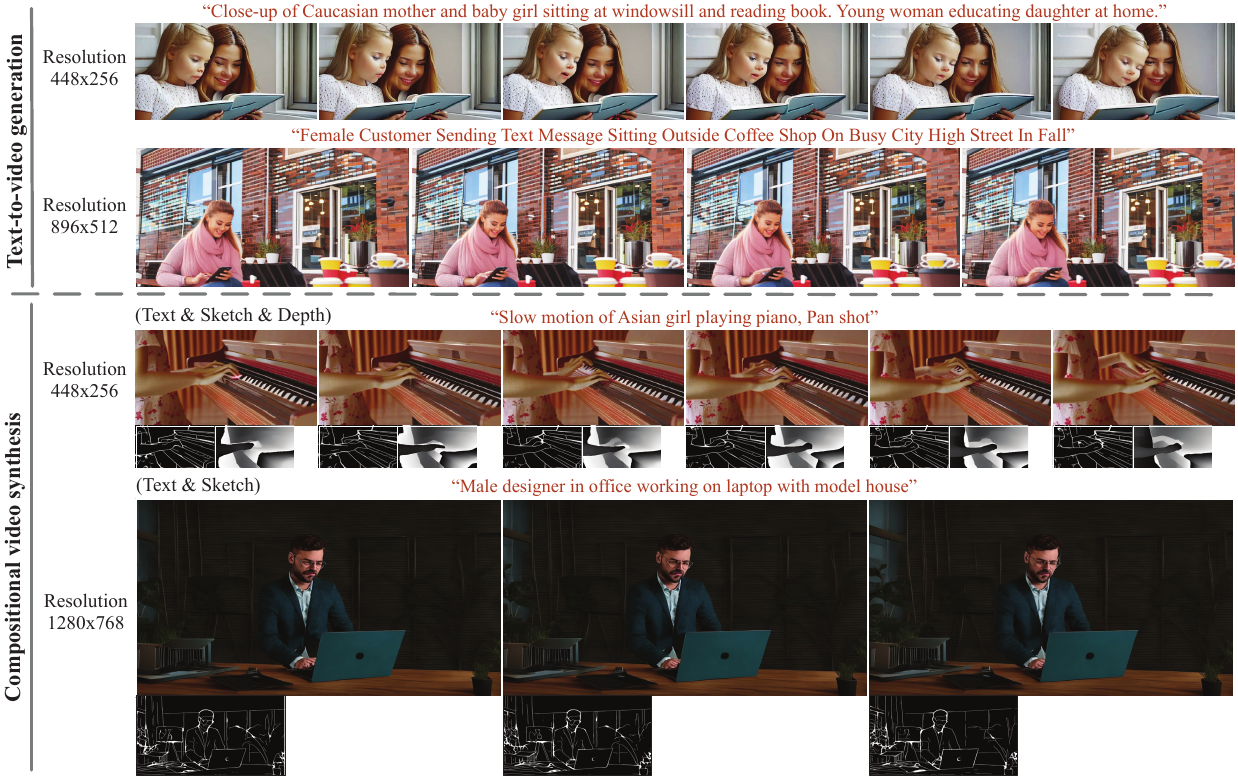}
    \vspace{-12pt}
    \captionof{figure}{
        \textbf{Example video results} generated by the proposed \method on text-to-video generation and compositional video synthesis tasks \emph{without training on any video-text pairs}.
        %
        % Since only language-free videos are required, High-definition videos can be easily generated.
        % High-definition videos can be 
        %
    }
    \label{first_figure}
    \vspace{15pt}
    }
]

\let\thefootnote\relax\footnotetext{$*$ Intern at Alibaba Group. }

% Regular Paper
\begin{abstract}

Diffusion-based text-to-video generation has witnessed impressive progress in the past year yet still falls behind text-to-image generation.
% \textit{v.s}
One of the key reasons is the limited scale of publicly available data (\textit{e.g.}, 10M video-text pairs in WebVid10M \textit{vs.} 5B image-text pairs in LAION), considering the high cost of video captioning.
Instead, it could be far easier to collect unlabeled clips from video platforms like YouTube.
Motivated by this, we come up with a novel text-to-video generation framework, termed \method, which can directly \textbf{learn with text-free videos}.
The rationale behind is to separate the process of text decoding from that of temporal modeling.
%
% To this end, we employ a content branch and a motion branch, which are jointly optimized with weights shared.
To this end, we employ a content branch and a motion branch, which are jointly optimized with weights shared.
Following such a pipeline, we study the effect of doubling the scale of training set (\textit{i.e.}, video-only WebVid10M) with some randomly collected text-free videos and are encouraged to observe the performance improvement (FID from 9.67 to 8.19 and FVD from 484 to 441), demonstrating the scalability of our approach.
We also find that our model could enjoy sustainable performance gain (FID from 8.19 to 7.64 and FVD from 441 to 366) after reintroducing some text labels for training.
Finally, we validate the effectiveness and generalizability of our ideology on both native text-to-video generation and compositional video synthesis paradigms.
% made public
Code and models will be publicly available  at \href{https://tf-t2v.github.io/}{here}.
% }

\end{abstract}
\vspace{-10pt}

% \vspace{-10pt}
\section{Introduction}
\label{sec:intro}

% widespread
Video generation aims to synthesize realistic videos that possess visually appealing spatial contents and temporally coherent motions.
% coherence.
% , enabling various applications such as video editing, creative production, and movie creation. 
%  remarkable， plausible
It has witnessed unprecedented progress in recent years with the advent of deep generative techniques~\cite{cogvideo,mocogan}, especially with the emergence of video diffusion models~\cite{modelscopet2v,tune-a-video,VideoLDM,zhang2023i2vgen,qing2023hierarchical,ma2023dreamtalk,wang2023videolcm}.
% \wx{
Pioneering approaches~\cite{Text2video-zero,tune-a-video,luo2023videofusion} utilize pure image diffusion models or fine-tuning on a small amount of video-text data to synthesize videos, leading to temporally discontinuous results due to insufficient motion perception~\cite{qi2023fatezero,zhang2023controlvideo}.
To achieve plausible results, current text-to-video methods like VideoLDM~\cite{VideoLDM} and ModelScopeT2V~\cite{modelscopet2v} usually insert temporal blocks into latent 2D-UNet~\cite{stablediffusion} and train the model on expansive video-text datasets, \eg, WebVid10M~\cite{webvid10m}.
To enable more controllable generation,
VideoComposer~\cite{videocomposer} proposes a compositional paradigm that incorporates additional conditions (\eg, depth, sketch, motion vectors, \etc.) 
to guide synthesis, allowing customizable creation.

Despite this, the progress in text-to-video generation still falls behind text-to-image generation~\cite{stablediffusion,Dalle2}.
One of the key reasons is the limited scale of publicly available video-text data, considering the high cost of video captioning~\cite{zhuo2023synthesizing}.
Instead, it could be far easier to collect text-free video clips from media platforms like YouTube.
% To leverage text-free videos,
% To reduce video-text annotations,
There are some works sharing similar inspiration,
% There also exists some works that 
% To this end,
Make-A-Video~\cite{make-a-video} and Gen-1~\cite{Gen-1} employ a two-step strategy that first leverages a large ($\sim$1B parameters) diffusion prior model~\cite{Dalle2} to convert text embedding into image embedding of CLIP~\cite{CLIP} and then enters it into an image-conditioned generator to synthesize videos.
% inference, some
However,
% Nevertheless,
the separate two-step manner may cause issues such as error accumulation~\cite{SceneScape}, increased model size and latency~\cite{Dalle2,xing2023simda}, and does not support text-conditional optimization if extra video-text data is available, leading to sub-optimal results.
% property
Moreover, the characteristics of scaling potential on video generation are still under-explored.
In this work, we aim to train a single unified video diffusion model that allows text-guided video generation by exploiting the widely accessible text-free videos and explore its scaling trend.
% or partially paired video-text data.
% }
%
% 
% To this end,
To achieve this,
we present a novel two-branch framework named \method, where a content branch is designed for spatial appearance generation, and a motion branch specializes in temporal dynamics synthesis.
More specifically, we utilize the publicly available image-text datasets such as LAION-5B~\cite{schuhmann2022laion} to learn text-guided and image-guided spatial appearance generation.
% 
%
% On the other hand,  explicit
In the motion branch, we harness the video-only data to conduct image-conditioned video synthesis, allowing the temporal modules to learn intricate motion patterns without relying on textual annotations.
%
% Video-text data if available can also be incorporated to optimize jointly.
% Video-text pairs
Paired video-text data, if available, can also be incorporated into co-optimization.
%
% Furthermore, in contrast to previous methods that impose training loss on each frame individually,  we introduce a temporal coherence loss to explicitly enforce the learning of correlations between adjacent frames, enhancing the continuity of generated videos.
Furthermore, unlike previous methods that impose training loss on each frame individually,  we introduce a temporal coherence loss to explicitly enforce the learning of correlations between adjacent frames, enhancing the continuity of generated videos.
%
% Note that, overcoming
% we study the scaling trend, multiple
In this way, the proposed \method achieves text-to-video generation by assembling contents and motions with a unified model, overcoming the high cost of video captioning and eliminating the need for complex cascading steps.

Notably, \method is a plug-and-play  paradigm, which can be integrated into existing text-to-video generation and compositional video synthesis frameworks as shown in \cref{first_figure}.
Different from most prior works that rely heavily on video-text data and train models on the widely-used watermarked and low-resolution (around 360P) WebVid10M~\cite{webvid10m},
% Experimental results on 
\method opens up new possibilities for optimizing with text-free videos or partially paired video-text data,
% training video diffusion models without the reliance on text annotations,  accessible, widespread
making it more scalable and versatile in widespread scenarios, such as high-definition video generation.
% properties, video-only WebVid10M
To study the scaling trend, we double the scale of the training set with some randomly collected text-free videos and are encouraged to observe the performance improvement, with FID from 9.67 to 8.19 and FVD from 484 to 441.
Extensive quantitative and qualitative experiments collectively demonstrate the effectiveness and scaling potential of the proposed \method in terms of synthetic continuity, fidelity, and controllability.

\section{Related Work}
\label{sec:related_work}

\begin{figure*}[t]
    \centering
\includegraphics[width=0.99\textwidth]{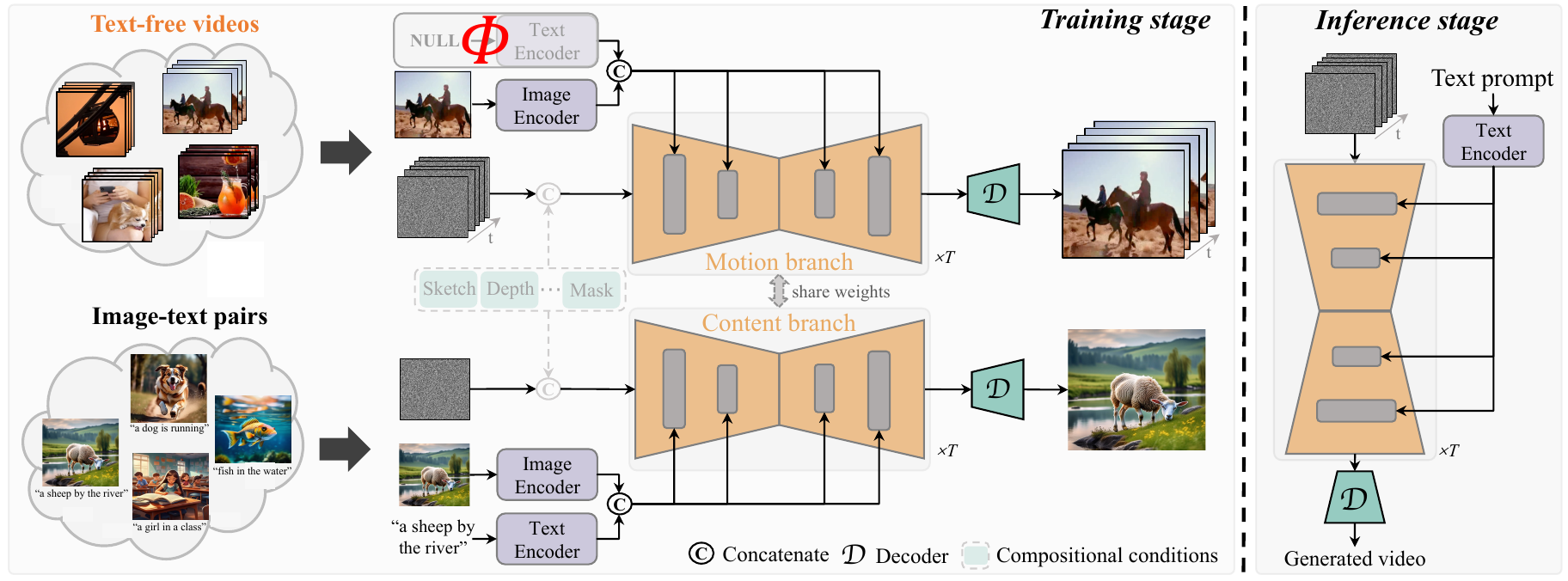}
\vspace{-4mm}
    \caption{
\textbf{Overall pipeline} of \method, which consists of two branches.
% In the training stage, The image branch utilizes paired image-text data to learn spatial appearance generation that is conditioned on both text and image.
In the content branch,
paired image-text data is leveraged to learn text-conditioned and image-conditioned spatial appearance generation.
% The video branch, on the other hand, supports the training of motion dynamic synthesis by feeding language-free videos or partially paired video-text data if available.
The motion branch supports the training of motion dynamic synthesis by feeding text-free videos (or partially paired video-text data if available).
% to train motion dynamic synthesis.
%under image conditions.
%
% For partially paired video-text data, 
%
During the training stage, both branches are optimized jointly.
% The above two branches are optimized jointly during the training stage.
%
Notably, \method can be seamlessly integrated into the compositional video synthesis framework by incorporating composable conditions.
In inference, \method enables text-guided video generation by taking text prompts and random noise sequences as input.
% is able to generating videos by taking text prompts and random noises as input to generate videos.
%
% An example of a subfigure.
}
    \label{Network}
    \vspace{-4mm}
\end{figure*}

% related work 
In this section, we provide a brief review of 
relevant literature on text-to-image generation, text-to-video generation, and compositional video synthesis.
% and diffusion models. controllable

\vspace{1mm}
\noindent \textbf{Text-to-image generation.}  
Recently, text-to-image generation has made significant strides with the development of large-scale image-text datasets such as LAION-5B~\cite{schuhmann2022laion}, allowing users to create high-resolution and photorealistic images that accurately depict the given natural language descriptions.
% Given natural language descriptions,  text-to-image generation aims to synthesize high-fidelity images with high alignment with the descriptions. 
% Previous methods mostly relied on the utilization of generative adversarial networks (GANs) for synthesis.
Previous methods~\cite{goodfellow2014generative,kang2023scaling,shen2021closed} primarily focus on synthesizing images by adopting generative adversarial networks (GANs) to estimate training sample distributions.
% demonstrate
Distinguished by the promising stability and scalability, diffusion-based generation models have attracted increasing attention~\cite{Dalle2,stablediffusion,kawar2023imagic,ruiz2023dreambooth,saharia2022photorealistic}.
Diffusion models utilize iterative steps to gradually refine the generated image, resulting in improved quality and realism.
Typically, Imagen~\cite{saharia2022photorealistic} and GLIDE~\cite{GLIDE} explore text-conditional diffusion models and boost sample quality by applying classifier-free guidance~\cite{ho2022classifierfreeguidance}.
DALL$\cdot$E 2~\cite{Dalle2} first leverages an image prior to bridge multi-modal embedding spaces and then learns a diffusion decoder to synthesize images in the pixel space.
Stable Diffusion~\cite{stablediffusion} introduces latent diffusion models that conduct iterative denoising processes at the latent level to save computational costs.
There are also some works that generate customized and desirable images by incorporating additional spatial control signals~\cite{huang2023composer,controlnet,T2i-adapter}.

\vspace{1mm}
\noindent \textbf{Text-to-video generation.}  
%
% Traditional 
% This task is more challenging compared to text-to-image generation as it requires capturing temporal dynamics and coherent motion.
% 
This task poses additional challenges compared to text-to-image generation due to the temporal dynamics involved in videos. 
% .
Various early techniques have been proposed to tackle this problem, such as recurrent neural networks combined with GANs~\cite{mocogan,wang2023styleinv,skorokhodov2022stylegan,balaji2019conditional,wang2020g3an} or transformer-based autoregressive models~\cite{yu2023magvit,cogvideo}. 
With the subsequent advent of video diffusion models pretrained on large-scale video-text datasets~\cite{webvid10m,HD-VILA-100M,wang2023internvid}, video content creation has demonstrated remarkable advances~\cite{VideoLDM,tune-a-video,wei2023dreamvideo,yuan2023instructvideo,ge2023preserve,he2022latent,ceylan2023pix2video,chai2023stablevideo,molad2023dreamix,yu2023video,luo2023videofusion,lu2023vdt,wang2023videofactory,geyer2023tokenflow,ni2023conditional,xing2023simda,wang2023zero,guo2023animatediff,chen2023motion,chen2023control,imagenvideo,Text2video-zero,liu2023video,an2023latent,qi2023fatezero,huang2023free,wang2023lavie,zhang2023show}.
% significantly limited due to the high computation
Imagen Video~\cite{imagenvideo} learns cascaded pixel-level diffusion models to produce high-resolution videos.
%
% Following DALL$\cdot$E 2
Following~\cite{Dalle2}, Make-A-Video~\cite{make-a-video} introduces a two-step strategy that first maps the input text to image embedding by a large ($\sim$1B parameters) diffusion prior model and then embeds the resulting embedding into an image-conditional video diffusion model to synthesize videos in pixel space.
% operate
VideoLDM~\cite{VideoLDM} and ModelScopeT2V~\cite{modelscopet2v} extend 2D-UNet into 3D-UNet by injecting temporal layers and operate a latent denoising process to save computational resources.
% Contrary to these approaches
In this paper, 
we present a single unified framework for text-to-video generation and study the scaling trend by harnessing widely accessible text-free videos.

% 

% \vspace{1mm}
\noindent \textbf{Compositional video synthesis.}  
Traditional text-to-video methods solely rely on textual descriptions to control the video generation process, limiting desired fine-grained customization such as texture, object position, motion patterns, \etc. To tackle this constraint and pursue higher controllability, several controllable video synthesis methods~\cite{videocomposer,Gen-1,zhang2023controlvideo,chen2023motion,chen2023control,yin2023dragnuwa,xing2023make,lapid2023gd,zhao2023controlvideo} have been proposed. These methods utilize additional control signals, such as depth or sketch, to guide the generation of videos. 
% finely, signals
By incorporating extra structured guidance, the generated content can be precisely controlled and customized.
% groundbreaking
Among these approaches, VideoComposer~\cite{videocomposer} stands out as a pioneering and versatile compositional technique. It integrates multiple conditioning signals including textual, spatial and temporal conditions within a unified framework, offering enhanced controllability, compositionality, and realism in the generated videos.
% enabling more comprehensive and customizable video generation.
% VideoComposer~\cite{videocomposer} is versatile 
Despite the remarkable quality, these methods still rely on high-quality video-text data to unleash powerful and customizable synthesis. 
%, limiting further evolution. 
% generation capabilities.
In contrast, our method can be
directly merged into existing controllable frameworks to customize videos by exploiting text-free videos.
% without training on any video-text pairs. 
% the cross-attention layers with much less computational overhead.

%
% 
% \section{Method}

% 
% We introduce 

\section{Method}
We first provide a brief introduction to the preliminaries of the video diffusion model.
Then, we will elaborate on the mechanisms of \method in detail.
The overall framework of the proposed \method is displayed in \cref{Network}.

\subsection{Preliminaries of video diffusion model}
% manner
% We give a brief introduction to video diffusion model.
% backward
Diffusion models involve a forward diffusion process and a reverse iterative denoising stage.
% Diffusion model gradually denoises the perturb data $x_{0}$ by performing $T$ consecutive forward processes:
The forward process of diffusion models  is gradually imposing random noise to clean data $x_{0}$  in a Markovian chain:
% the perturb data $x_{0}$ by performing $T$ consecutive forward processes:
\begin{equation}
  q(x_{t}|x_{t-1}) = \mathcal{N}(x_{t};\sqrt{1-\beta_{t-1}}x_{t-1}, \beta_{t}I), t=1,...,T
  \label{eq:1}
\end{equation}
where $\beta_{t} \in (0,1)$ is a noise schedule and $T$ is the total time step.  
% variance
When $T$ is sufficiently large, \eg $T=1000$, the resulting $x_{T}$ is nearly a random Gaussian distribution $\mathcal{N}(0,I)$.
% estimate  as follows
The role of diffusion model is to denoise $x_{T}$ and learn to iteratively estimate the reversed process:
\begin{equation}
p_{\theta }(x_{t-1}|x_{t}) = \mathcal{N}(x_{t-1};\mu_{\theta}(x_{t},t),{\textstyle \sum_{\theta}}(x_{t},t) )
  \label{eq:2}
\end{equation}
%
% In text-to-video generation,
We usually train a denoising model $\hat{x}_{\theta}$ parameterized by $\theta$ to approximate the original data $x_{0}$ and optimize the following v-prediction~\cite{vprediction,imagenvideo} problem:
\begin{equation}
\mathcal{L}_{base} = \mathbb{E}_{\theta}[|| v - \hat{x}_{\theta}(x_{t},t,c) ||_{2}^{2}]
  \label{eq:3}
\end{equation}
where $c$ is conditional information such as textual prompt, and $v$ is the parameterized prediction objective. 
% In existing video diffusion model
In representative video diffusion models~\cite{VideoLDM,modelscopet2v,videocomposer}, 
% the process is usually converted to the latent space by applying a variational autoencoder~\cite{stablediffusion},
% and
the denoising model $\hat{x}_{\theta}$ is a latent 3D-UNet~\cite{VideoLDM,modelscopet2v} modified from its 2D version~\cite{stablediffusion} by inserting additional temporal blocks, which is optimized in the latent feature space by applying a variational autoencoder~\cite{esser2021taming}, and \cref{eq:3} is applied on each frame of the input video to train the whole model.
% 

% .

% FreeVideoDiff
% \subsection{FreeVDM}
\subsection{TF-T2V}

The objective of \method is to learn a text-conditioned video diffusion model to create visually appealing and temporally coherent videos with text-free videos or partially paired video-text data.
Without loss of generality, we first describe the workflow of our \method in the scenario where only text-free video is used.
With merely text-free videos available for training, it is challenging to guide content creation by textual information since there lacks text-visual correspondence.
% between textual descriptions and visual content,. 
% text-guided model by introduce
% To this end, 
To tackle this issue,
we propose to resort to web-scale and high-quality image-text datasets~\cite{Laion-400m,schuhmann2022laion}, which are publicly accessible on the Internet.
% another, an additional
However, this raises another question: \emph{how can we leverage the image-text data and text-free videos in a unified framework?} 
%

% 视频表观和图片表观还是有差异的
Recalling the network architecture in 3D-UNet, the spatial modules mainly focus on appearance modeling, and the temporal modules primarily aim to operate motion coherence.
The intuition is that we can utilize image-text data to learn text-conditioned spatial appearance generation and adopt high-quality text-free videos to guide consistent motion dynamic synthesis.
% , enabling text-to-video generation in a single model.  
In this way, we can perform text-to-video generation in a single model to synthesize high-quality and consistent videos during the inference stage.
Based on this, the proposed \method consists of two branches: a content branch for spatial appearance generation and a motion branch for motion dynamic synthesis.
% Videos contains 

% 

% \noindent
\subsubsection{Spatial appearance generation}

% \ind
% \textbf{Spatial appearance generation.}
% To learn spatial appearance generation, 
Like previous text-to-image works~\cite{stablediffusion,controlnet}, the content branch of \method takes a noised image $I_{image} \in H \times W \times C$ as input, where $H$, $W$, $C$ are the height, width, and channel dimensions respectively, and 
employs conditional signals (\ie, text and image embeddings) to offer semantic guidance for content generation.
% primarily, prevailingly, chiefly
This branch primarily concentrates on optimizing the spatial modules in the video diffusion model and plays a crucial role in determining appealing visual quality.
% is expected to . concerning
% The video diffusion model  
% control the semantic 
In order to ensure that each condition can also control the created content separately, we randomly drop text or image embeddings with a certain probability during training.
The text and image encoders from CLIP~\cite{CLIP} are adopted to encode embeddings.

\begin{table*}[t]
    % \tablestyle{4pt}{0.95}
    \caption{
        \textbf{Quantitative comparison} with state-of-the-art methods for text-to-video task on MSR-VTT in terms of FID, FVD, and CLIPSIM. 
    }
    \label{tab:compare_with_sota_t2v}
    \vspace{-8pt}
    \centering
    \tablestyle{4pt}{0.95}
    \setlength{\tabcolsep}{12pt}
   % \begin{tabular}{@{}lc@{}}
   % \resizebox{\textwidth}{20mm}{
   % \scalebox{0.98}{
      \begin{tabular}{l|ccccc}
    % \shline
\hspace{-3mm}        Method      &  \hspace{1mm} Zero-shot \hspace{1mm}  & Parameters & \hspace{1mm} FID ($\downarrow$)\hspace{1mm} &  \hspace{1mm} FVD  ($\downarrow$) \hspace{1mm} & CLIPSIM ($\uparrow$) \\
        \shline
\hspace{-3mm}         N{\"u}wa~\cite{wu2022nuwa}    & No & - & 47.68 & - & 0.2439 \\
\hspace{-3mm}         CogVideo (Chinese)~\cite{cogvideo}    & Yes & 15.5B & 24.78 & - & 0.2614 \\
\hspace{-3mm}         CogVideo (English)~\cite{cogvideo}    & Yes & 15.5B & 23.59 & 1294 & 0.2631 \\
\hspace{-3mm}         MagicVideo~\cite{zhou2022magicvideo}    & Yes & - & - & 1290 & - \\
\hspace{-3mm}         Make-A-Video~\cite{make-a-video}    & Yes & 9.7B & 13.17 & - & \textbf{0.3049} \\
\hspace{-3mm}         ModelScopeT2V~\cite{modelscopet2v}    & Yes & 1.7B & 11.09 & 550 & 0.2930 \\
\hspace{-3mm}         VideoComposer~\cite{videocomposer}    & Yes & 1.9B & 10.77 & 580 & 0.2932 \\ %  10.77
\hspace{-3mm}         Latent-Shift~\cite{an2023latent} & Yes & 1.5B & 15.23 & - & 0.2773 \\
\hspace{-3mm}         VideoLDM~\cite{VideoLDM}    & Yes & 4.2B & - & - & 0.2929 \\
\hspace{-3mm}         PYoCo~\cite{ge2023preserve}    & Yes & - & 9.73 & - & - \\
        \hline
        \rowcolor{Gray}
\hspace{-3mm}         \method (WebVid10M)   &  Yes & 1.8B & {9.67}  & {484}  & 0.2953  \\
        \rowcolor{Gray}
\hspace{-3mm}         \method (WebVid10M+Internal10M)  &  Yes & 1.8B & \textbf{8.19}  & \textbf{441}  & 0.2991  \\
        % \shline
       % \shline
    \end{tabular}
% \end{wraptable}
    \vspace{-10pt}
\end{table*}

\subsubsection{Motion dynamic synthesis}

% \textbf{Motion dynamic synthesis.}
% principle, standard
The pursuit of producing highly temporally consistent videos is a unique hallmark of video creation.
Recent advancements~\cite{modelscopet2v,VideoLDM,videocomposer,wang2023videofactory} in the realm of video synthesis usually utilize large-scale video-text datasets such as WebVid10M~\cite{webvid10m} to achieve coherent video generation.
However, acquiring large-scale video-text pairs consumes extensive manpower and time, hindering the scaling up of video diffusion models.
% to make matters worse, the widely used WebVid10M is a watermarked and low-resolution data set, resulting in unsatisfactory video generation that cannot meet today's high-quality video generation requirements.
To make matters worse, the widely used WebVid10M is a watermarked and low-resolution (around 360P) dataset, resulting in unsatisfactory video creation that cannot meet the high-quality video synthesis requirements.

To mitigate the above issues,
% To this end, 
we propose to leverage high-quality text-free videos that are easily accessible on video media platforms, \eg, YouTube and TikTok.
% potential
To fully excavate the abundant motion dynamics within the text-free videos, we train a image-conditioned model.
By optimizing this image-to-video generation task, the temporal modules in the video diffusion model can learn to perceive and model diverse motion dynamics.
Specifically,
given a noised video $I_{video} \in F\times H \times W \times C $, where $F$ is the temporal length, the motion branch of \method learns to recover the undisturbed video guided by the image embedding.
The image embedding is extracted from the center frame of the original video by applying CLIP's image encoder~\cite{CLIP}.
%

% massive, spacious, extensive
Since large-scale image-text data used for training contains abundant movement intentions~\cite{li2022action}, \method can achieve text-to-video generation by assembling spatial appearances involving  motion trends and predicted motion dynamics.
% By decomposing the video generation process into spatial appearance generation and temporal dynamics synthesis, \method can effectively learn the spatial dependencies and temporal dynamics within a unified model.
% additional precise
When extra paired video-text data is available, we conduct both text-to-video and image-to-video generation based on video-text pairs to train \method and further enhance the perception ability for desirable textual control.
%
% By optimizing image-to-video generation, the temporal modules in video diffusion model can learn to perceive and model diverse motion dynamics.

% 

% 
%
In addition, we notice that
previous works apply the training loss (\ie, \cref{eq:3}) on each frame of the input video individually without considering temporal correlations between frames, suffering from
 incoherent appearances and motions.
 % inconsistency of geometry and motion.
% Different from previous works that apply the training loss (\ie, \cref{eq:3}) on each frame of the input video individually, the proposed loss term measures the discrepancy between the predicted frame differences and the ground truth frame differences of the input parameterized video:
Inspired by the early study~\cite{wang2021tdn,zhang2012slow,wang2023molo,jhuang2013towards} finding that the difference between two adjacent frames usually contains motion patterns, \eg, dynamic trajectory, we thus propose a temporal coherence loss that utilizes the frame difference as an additional supervisory signal:
%
% The proposed loss term measures the discrepancy between the predicted frame differences and the ground truth frame differences of the input parameterized video:
\begin{equation}
\mathcal{L}_{coherence} = \mathbb{E}_{\theta}[{\textstyle \sum_{j=1}^{F-1}} || (v_{j+1}-v_{j}) - (o_{j+1}-o_{j}) ||_{2}^{2}]
\label{eq:4}
\end{equation}
%
% where $o_{j}$ and $v_{j}$ are the predicted parameterized frame and corresponding ground truth objective.
where $o_{j}$ and $v_{j}$ are the predicted frame and corresponding ground truth.
% measures
This loss term measures the discrepancy between the predicted frame differences and the ground truth frame differences of the input parameterized video.
%
% Since frame differences contain motion information, minimizing
% \cref{eq:4} can explicitly enforce smoothness between adjacent video frames.
%
% By enforcing temporal coherence, 
By minimizing \cref{eq:4},
\method helps to alleviate frame flickering and ensures that the generated videos exhibit seamless transitions and promising temporal dynamics.

\subsubsection{Training and inference}
% \textbf{Inference}

% To eliminate the domain gap between image and video and 
In order to mine the complementary advantages of spatial appearance generation and motion dynamic synthesis, we jointly optimize the entire model in an end-to-end manner.
The total loss can be formulated as:
\begin{equation}
\mathcal{L}_{total} = \mathcal{L}_{base} + \lambda \mathcal{L}_{coherence}
\label{eq:5}
\end{equation}
where $\mathcal{L}_{base}$ is imposed on video and image together by treating the image as a ``single frame" video, and $\lambda$ is a balance coefficient that is set empirically to 0.1.
% , and $\mathcal{L}_{coherence}$ applies only to video branch.
% The end-to-end
% In the case that par

After training, we can perform text-guided video generation to synthesize temporally consistent video content that aligns well with the given text prompt.
%
% In addition,
Moreover, \method is a general framework and can also be inserted into existing compositional video synthesis paradigm~\cite{videocomposer} by incorporating additional spatial and temporal structural conditions, allowing for customized video creation.

\section{Experiments}

% In this section, we present a comprehensive quantitative and qualitative evaluation of the proposed \method for text-to-video generation and composition video synthesis.
In this section, we present a comprehensive quantitative and qualitative evaluation of the proposed \method on text-to-video generation and composition video synthesis.
% in terms of quantitative and qualitative evaluation. 
% 

% \subsection{Datasets and experimental setup}
\subsection{Experimental setup}

% \noindent \textbf{dataset}

\noindent \textbf{Implementation details.}
\method is built on two typical open-source baselines, \ie, ModelScopeT2V~\cite{modelscopet2v} and VideoComposer~\cite{videocomposer}.
%
% Following~\cite{videocomposer,modelscopet2v}, 
DDPM sampler~\cite{DDPM} with $T=1000$ steps is adopted for training, and we employ DDIM~\cite{DDIM} with 50 steps for inference.
% by default. 
%
We optimize \method using AdamW optimizer with a learning rate of 5e-5.
%
% To speed up the training process, Stable Diffusion v2.1~\cite{stablediffusion} is utilized as weight initialization.
%
% The whole diffusion process is conducted at the latent space, and  compression factor of the variational autoencoder is 8.
%
For input videos, we sample 16 frames from each video at 4FPS and crop a $448\times 256$ region at the center as the basic setting. 
Note that we can also easily train high-definition video diffusion models by collecting high-quality text-free videos (see examples in the Appendix).
% For high-definition video generation, we 
LAION-5B~\cite{schuhmann2022laion} is utilized to provide image-text pairs.
%
% Unless otherwise stated, we treat WebVid10M that includes about 10.7M video-text pairs as a text-free dataset to train \method and do not use any textual annotations.
Unless otherwise stated, we treat WebVid10M, which includes about 10.7M video-text pairs, as a text-free dataset to train \method and do not use any textual annotations.
To study scaling trends, we gathered about 10M high-quality videos without text labels from internal data, termed the Internal10M dataset.
% from private data.
% 

\begin{figure*}[t]
    \centering
\includegraphics[width=0.97\textwidth]{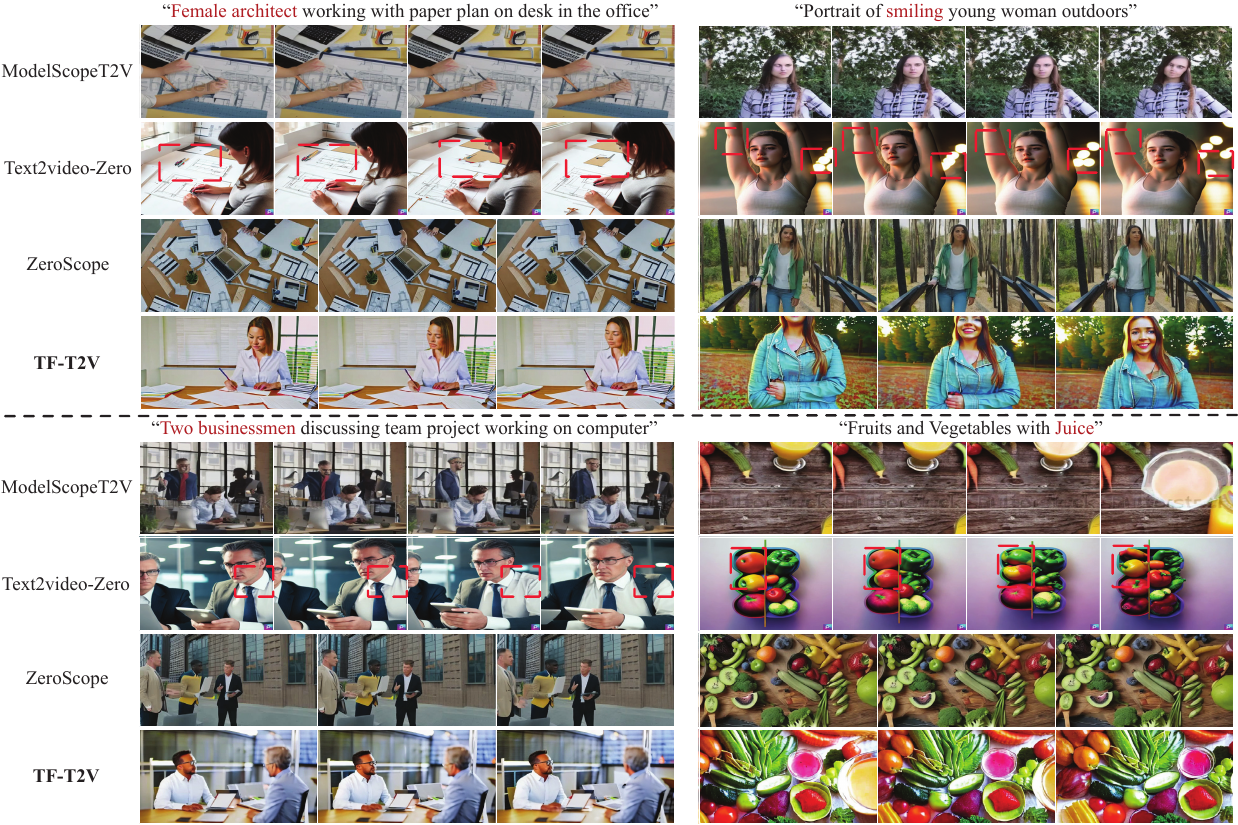}
\vspace{-4mm}
% existing
    \caption{\textbf{{Qualitative comparison on text-to-video generation}}. Three representative open-source text-to-video approaches are compared, including ModelScopeT2V~\cite{modelscopet2v}, Text2video-Zero~\cite{Text2video-zero} and ZeroScope~\cite{ZeroScope}. 
    % Appendix, supplementary materials
    Please refer to the Appendix for videos and more comparisons.
    }
    \label{fig:Compare_T2V}
    \vspace{-4mm}
\end{figure*}

\vspace{1mm}
\noindent \textbf{Metrics.}
(i) To evaluate text-to-video generation,
following previous works~\cite{VideoLDM,modelscopet2v}, we leverage the standard Fr{\'e}chet Inception Distance (FID),  Fr{\'e}chet Video Distance (FVD), and CLIP Similarity (CLIPSIM) as quantitative
evaluation metrics and report results on MSR-VTT dataset~\cite{xu2016msr}.
(ii) For controllability evaluation, we leverage depth error, sketch error, and end-point-error (EPE)~\cite{dosovitskiy2015flownet} to verify whether the generated videos obey the control of input conditions. 
% For instance, depth error measures the divergence between the input depth conditions and eliminated depth of the synthesised video.
Depth error measures the divergence between the input depth conditions and the eliminated depth  of the synthesized video.
% Likewise
Similarly, sketch error examines the sketch control.
% Sketch error  is similar to depth error.
%
EPE evaluates the flow consistency between the reference video and the generated video.
%
% In addition to the above metrics, we also introduce human evaluation to validate our method.
In addition, human evaluation is also introduced to validate our method.

\begin{table}[t]
    % \vspace{-1.1em}
    % \vspace{-3mm}
    \caption{
        {\textbf{Human preference results} on text-to-video generation.
    }}
    \label{tab:t2v_human}
    \vspace{-3mm}
    \tablestyle{4pt}{0.95}
    % \vspace{-5pt}
    \centering
    \setlength{\tabcolsep}{6pt}{
   % \begin{tabular}{@{}lc@{}}
   % \scalebox{0.98}{
      \begin{tabular}{l|ccc}
    % \shline
       \multirow{2}{*}{ Method} &  Text & Visual  & Temporal \\
                                &  alignment & quality & coherence \\
        \shline
        ModelScopeT2V~\cite{modelscopet2v}    &  83.5\% & 74.0\%  & 81.3\% \\
         \rowcolor{Gray}
        \method   &  \textbf{86.5\%} & \textbf{87.0\%} & \textbf{92.5\%} \\
        % \shline
       % \shline
 %      
%
    \end{tabular}
    }
    % \vspace{-10pt}
    \vspace{-4mm}
% \end{wraptable}
\end{table}

\subsection{Evaluation on text-to-video generation}

\cref{tab:compare_with_sota_t2v} displays the comparative quantitative results with existing state-of-the-art methods.
We observe that \method achieves remarkable performance under various metrics.
% strong scalable
Notably, \method trained on WebVid10M and Internal10M obtains higher performance than the counterpart on WebVid10M, revealing promising scalable capability.
% of our method.
%
We show the qualitative visualizations in \cref{fig:Compare_T2V}.
From the results, we can find that
compared with previous methods, \method obtains impressive video creation in terms of both temporal continuity and visual quality.
% \method obtains temporally consistent and visually appealing 
% A consistent conclusion is also revealed in the human assessment in Table 1
% Human assessment in \cref{tab:t2v_human} 
The human assessment in \cref{tab:t2v_human} also reveals the above observations.
% , with values 
The user study is performed on 100 randomly synthesized videos.
% The results are obtained from 100 randomly synthesized videos.

\subsection{Evaluation on compositional video synthesis}

\begin{table}[t]
    % \vspace{-1.1em} sequence
    \caption{
        {\textbf{Evaluation of structure control} based on depth signals.
    }}
    \label{tab:depth_error}
    \vspace{-3mm}
    \tablestyle{4pt}{0.95}
    % \vspace{-5pt}
    \centering
    \setlength{\tabcolsep}{7pt}{
   % \begin{tabular}{@{}lc@{}}
   % \scalebox{0.98}{
      \begin{tabular}{l|c|c}
    % \shline
        Method    & Condition  &  Depth error ($\downarrow$) \\
        \shline
        VideoComposer~\cite{videocomposer} & Text    &  0.382  \\
        VideoComposer~\cite{videocomposer} & Text and depth    &  0.217  \\
         \rowcolor{Gray}
         % \hline
        \method & Text and depth  &  \textbf{0.209} \\
        % \shline
       % \shline
 %      
%
    \end{tabular}
    }
    % \vspace{-10pt}
    \vspace{-2mm}
% \end{wraptable}
\end{table}

\begin{table}[t]
    % \vspace{-1.1em}
    \caption{
        {\textbf{Evaluation of structure control} based on sketch signals.
    }}
    \vspace{-3mm}
    \label{tab:sketch_error}
    \tablestyle{4pt}{0.95}
    % \vspace{-5pt}
    \centering
    \setlength{\tabcolsep}{7pt}{
   % \begin{tabular}{@{}lc@{}}
   % \scalebox{0.98}{
      \begin{tabular}{l|c|c}
    % \shline
        Method    & Condition  &  Sketch error ($\downarrow$) \\
        \shline
        VideoComposer~\cite{videocomposer} & Text    &  0.1854  \\
        VideoComposer~\cite{videocomposer} & Text and sketch    &  0.1161  \\
         \rowcolor{Gray}
         % \hline
        \method & Text and sketch  &  \textbf{0.1146} \\
        % \shline
       % \shline
 %      
%
    \end{tabular}
    }
    % \vspace{-10pt}
    \vspace{-2mm}
% \end{wraptable}
\end{table}

\begin{table}[t]
    % \vspace{-1.1em}
    % \vspace{-3mm}
    \caption{
        {\textbf{Evaluation of motion control} based on motion vectors.
    }}
    \label{tab:mv_error}
    \vspace{-3mm}
    \tablestyle{4pt}{0.95}
    % \vspace{-5pt}
    \centering
    \setlength{\tabcolsep}{7pt}{
   % \begin{tabular}{@{}lc@{}}
   % \scalebox{0.98}{
      \begin{tabular}{l|c|c}
    % \shline
        Method    & Condition  &  EPE ($\downarrow$) \\
        \shline
        VideoComposer~\cite{videocomposer} & Text    &  4.13  \\
        VideoComposer~\cite{videocomposer} & Text and motion vector    &  1.98  \\
         \rowcolor{Gray}
         % \hline
        \method & Text and motion vector  &  \textbf{1.88} \\
        % \shline
       % \shline
 %      
%
    \end{tabular}
    }
    % \vspace{-10pt}
    \vspace{-2mm}
% \end{wraptable}
\end{table}

\begin{figure*}[t]
    \centering
\includegraphics[width=0.99\textwidth]{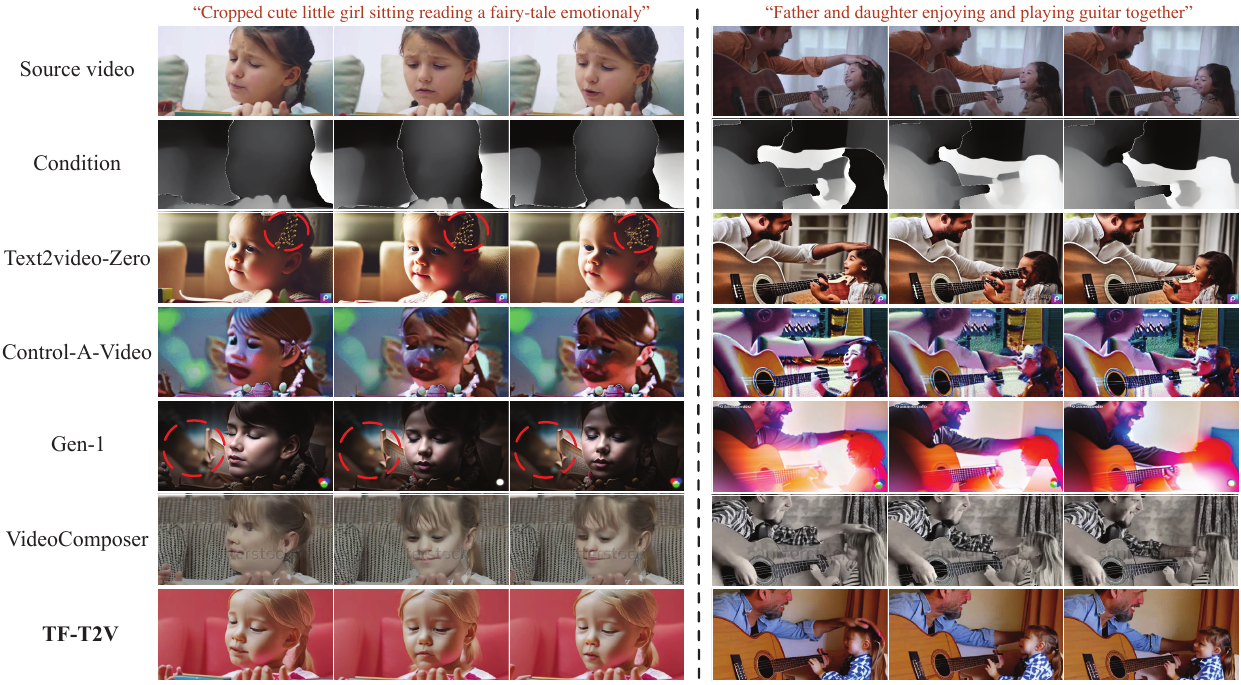}
\vspace{-4mm}
% descriptions
    \caption{\textbf{Qualitative comparison on compositional depth-to-video generation}. The videos are generated by taking textual prompts
and structural guidance as conditions. Compared with existing methods, \method yields more structural compliance and high-fidelity results.
}
    \label{fig:depth_compare}
    \vspace{-4mm}
\end{figure*}

\begin{figure*}[t]
    \centering
\includegraphics[width=0.97\textwidth]{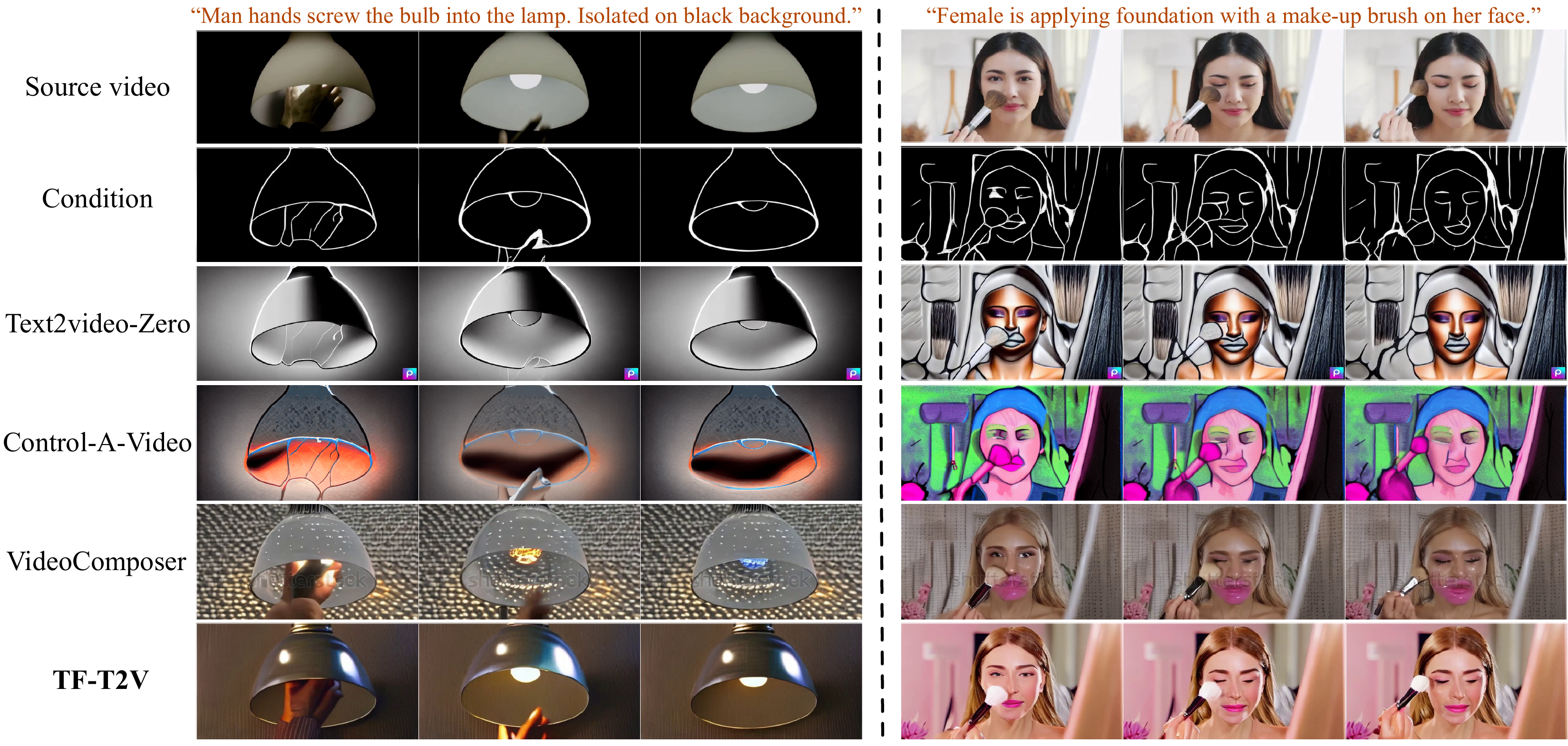}
\vspace{-4mm}
    \caption{\textbf{Qualitative comparison on compositional sketch-to-video generation}. The videos are generated by taking textual descriptions and structural guidance as conditions. 
    Compared with other methods, \method produces more realistic and consistent results.
    }
    \label{fig:Sketch_compare}
    \vspace{-4mm}
\end{figure*}

\begin{table}[t]
    % \vspace{-1.1em}
    \caption{
        {\textbf{Human evaluations} on compositional video synthesis.
    }}
    \label{tab:compositional_human}
    \vspace{-3mm}
    \tablestyle{4pt}{0.95}
    % \vspace{-5pt}
    \centering
    \setlength{\tabcolsep}{6.5pt}{
   % \begin{tabular}{@{}lc@{}}
   % \scalebox{0.98}{
      \begin{tabular}{l|ccc}
    % \shline
       \multirow{2}{*}{ Method} &  Structure & Visual  & Temporal \\
                                &  alignment & quality & coherence \\
        \shline
        VideoComposer~\cite{videocomposer}    &  79.0\% & 66.0\%  & 77.5\% \\
         \rowcolor{Gray}
         % \hline
        \method   &  \textbf{89.0\%} & \textbf{79.5\%} & \textbf{84.5\%} \\
        % \shline
       % \shline
 %      
%
    \end{tabular}
    }
    % \vspace{-10pt}
    \vspace{-2mm}
% \end{wraptable}
\end{table}

% We compare the structural and motion controllability of \method and VideoComposer in \cref{tab:depth_error,tab:sketch_error,tab:mv_error}.
We compare the controllability of \method and VideoComposer on 1,000 generated videos in terms of depth control (\cref{tab:depth_error}), sketch control (\cref{tab:sketch_error}) and motion control (\cref{tab:mv_error}).
The above experimental evaluations highlight the effectiveness of \method by leveraging text-free videos.
% exhibition
In \cref{fig:depth_compare} and \ref{fig:Sketch_compare}, we show the comparison of \method and existing methods on compositional video generation.
We notice that \method exhibits high-fidelity and consistent video generation.
% under the input criteria.
%
In addition, we conduct a human evaluation on 100 randomly sampled videos and report the results in \cref{tab:compositional_human}.
The preference assessment provides further evidence of the superiority of the proposed \method.

% Overall, our experimental evaluations highlight the effectiveness of our proposed method in decomposing video generation into spatial appearance and temporal dynamics synthesis. Moreover, our method showcases its ability to extend to compositional video generation, providing further evidence of its effectiveness. These results collectively emphasize the superiority of our approach over existing methods that heavily rely on video-text pairs.

\subsection{Ablation study}

\begin{figure*}[t]
    \centering
\includegraphics[width=0.99\textwidth]{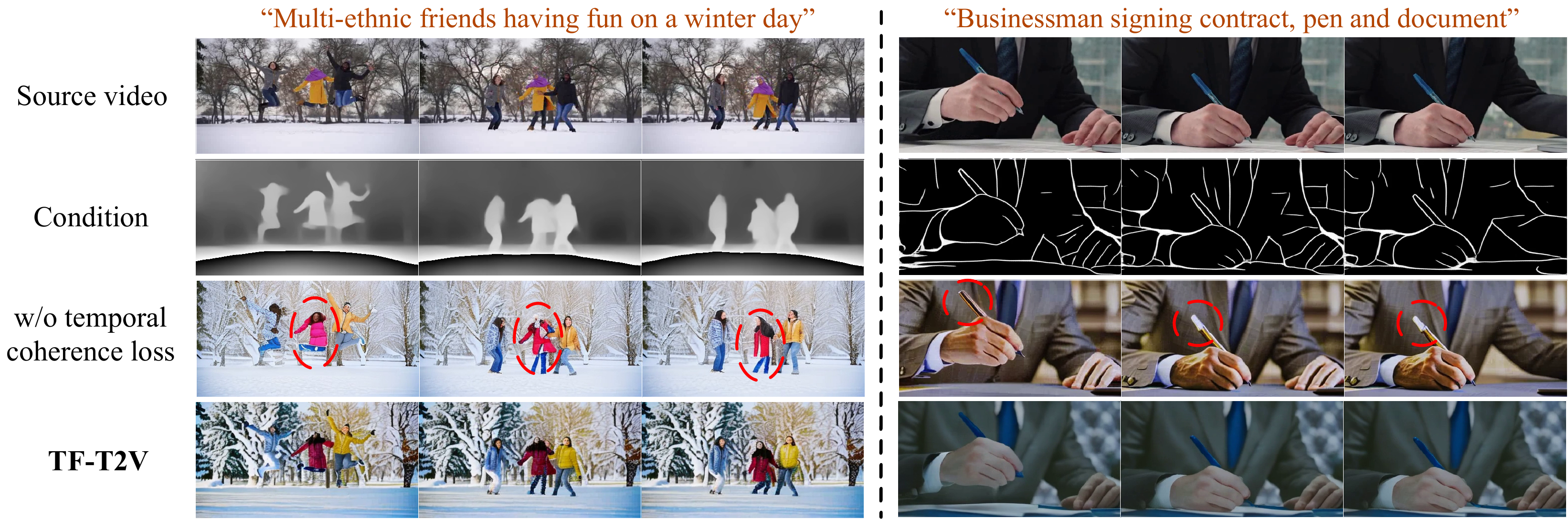}
\vspace{-3mm}
    \caption{
    \textbf{Qualitative ablation study}. The videos are generated by taking textual descriptions
and structural guidance as conditions.
%
% Without the temporal coherence loss, the synthesized videos suffer from temporal inconsistency such as color shift.
    %
    }
    \vspace{-2mm}
    \label{fig:ablation_study}
\end{figure*}

\noindent 
\textbf{Effect of temporal coherence loss.}
% the temporal consistency
To enhance temporal consistency, we propose a temporal coherence loss.
In \cref{tab:frame_consistency}, we show the effectiveness of the proposed temporal coherence loss in terms of frame consistency.
% color shift, values
The metric results are obtained by calculating the average CLIP similarity of two consecutive frames in 1,000 videos.
%
% Experimental results 
%
We further display the qualitative comparative results in \cref{fig:ablation_study} and observe that temporal coherence loss helps to alleviate temporal discontinuity such as color shift.
%
% Refer to 
% Please refer to the Appendix for more ablation studies.

\begin{table}[t]
    % \vspace{-1.1em}
    \caption{
        {\textbf{Text-to-video evaluation} on frame consistency.
    }}
    \label{tab:frame_consistency}
    \vspace{-3mm}
    \tablestyle{4pt}{0.95}
    % \vspace{-5pt}
    \centering
    \setlength{\tabcolsep}{8pt}{
   % \begin{tabular}{@{}lc@{}}
   % \scalebox{0.98}{
      \begin{tabular}{l|c}
    % \shline
        Method      &  Frame consistency (\%) $\uparrow$ \\
        \shline
        \textit{w/o} temporal coherence loss    &  89.71 \\
         \rowcolor{Gray}
        \method   &  \textbf{91.06} \\
        % \shline
       % \shline
 %      
%
    \end{tabular}
    }
    % \vspace{-10pt}
    \vspace{-3mm}
% \end{wraptable}
\end{table}

\begin{table}[t]
    % \vspace{-1.1em}
    \caption{
        {
        % comparision
        \textbf{Quantitative experiments on text-to-video generation}. % comparison
        % ModelScopeT2V uses labeled WebVid10M to train the model. In \method, only language-free videos are leveraged.
        \method-Semi means the semi-supervised setting where labeled WebVid10M and text-free Internal10M are adopted. 
    }}
    \label{tab:compare_with_semi_t2v}
    \vspace{-2mm}
    \tablestyle{4pt}{0.95}
    % \vspace{-5pt}
    \centering
    \setlength{\tabcolsep}{6pt}{
   % \begin{tabular}{@{}lc@{}}
   % \resizebox{\textwidth}{20mm}{
   % \scalebox{0.98}{
      \begin{tabular}{l|ccc}
    % \shline
        Method       &  FID ($\downarrow$) &   FVD  ($\downarrow$)  & CLIPSIM ($\uparrow$) \\
        \shline
        ModelScopeT2V~\cite{modelscopet2v}   &  11.09 & 550 & 0.2930 \\
        \rowcolor{Gray}
        \method  &   {8.19}  & {441}  & {0.2991}  \\
        \rowcolor{Gray}
        \method-\emph{Semi}  &   \textbf{7.64}  & \textbf{366}  & \textbf{0.3032}  \\
        % \shline
       % \shline
 %      
%
    \end{tabular}
    }
    % \vspace{-10pt}
    \vspace{-3mm}
% \end{wraptable}
\end{table}

\begin{figure*}[t]
    \centering
\includegraphics[width=0.99\textwidth]{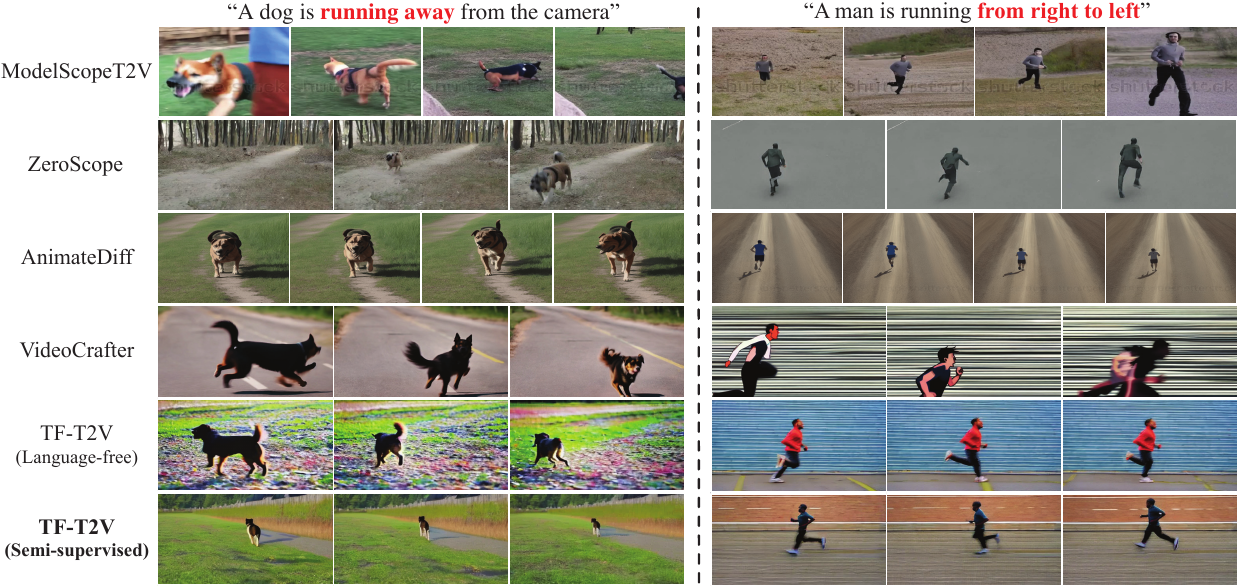}
\vspace{-4mm}
    \caption{
    \textbf{Qualitative evaluation} on text-to-video generation with temporally-correlated text prompts involving the evolution of 
movement.
    }
    \label{fig:Semi-supervised_t2v}
    \vspace{-3mm}
\end{figure*}

% % separately
% \noindent 
% \textbf{Effect of joint training.}
% % and eliminate the domain gap
% In our default setting, we jointly train the spatial and temporal blocks in the video diffusion model to fully exploit the complementarity between image and video modalities.
% %
% An alternative strategy is to separate the spatial and temporal modeling into sequential two stages.
% % temporal 
% We conduct a comparative experiment on these two strategies in \cref{tab:compare_with_train_manner}.
% The results demonstrate the rationality of joint optimization.

% semi-paired setting, semi-supervised setting
\subsection{Evaluation on semi-supervised setting}

Through the above experiments and observations, we verify that text-free video can help improve the continuity and quality of generated video.
%
% When partially paired video-text data is available, 
As previously stated,
\method also supports the combination of annotated video-text data and text-free videos to train the model, \ie, the semi-supervised manner.
% explicit, 
The annotated text can provide additional fine-grained motion signals, enhancing the alignment of generated videos and the provided prompts involving desired motion evolution.
We show the comparison results in \cref{tab:compare_with_semi_t2v} and find that the semi-supervised manner reaches the best performance, indicating the effectiveness of harnessing text-free videos.
% , demonstrating the .
Notably, \method-Semi outperforms ModelScopeT2V trained on labeled WebVid10M, possessing good scalability.
% properties
% 
%
%
% prove
Moreover, the qualitative evaluations in \cref{fig:Semi-supervised_t2v} show that existing methods may struggle to synthesize text-aligned consistent videos when textual prompts involve desired temporal evolution. 
% and suffer from inconsistent ability
In contrast, \method in the semi-supervised setting exhibits excellent text-video alignment and temporally smooth generation.
% uncover
% 

\section{Conclusion}

In this paper, we present a novel and versatile video generation framework named \method to exploit text-free videos and explore its scaling trend.
\method effectively decomposes video generation into spatial appearance generation and motion dynamic synthesis.
A temporal coherence loss is introduced to explicitly constrain the learning of correlations between adjacent frames.
Experimental results demonstrate the effectiveness and potential of \method in terms of fidelity, controllability, and scalability.
% generating high-quality consistent videos.

% 

\vspace{2mm}
\noindent{\textbf{Acknowledgements.}}
This work is supported by the National Natural Science Foundation
of China under grant U22B2053 and Alibaba Group through Alibaba Research Intern Program.

\vspace{-1mm}
{
\small
\bibliographystyle{ieeenat_fullname}
\bibliography{ref.bib}
}

% Supplementary Material
% \input{sec/X_suppl.tex}
\clearpage
% \setcounter{page}{1}
% \maketitlesupplementary
% \begin{figure*}[ht]
% % \twocolumn[
% {
% % \maketitle
% % \maketitlesupplementary
%     \centering
% \includegraphics[width=0.99\textwidth]{Picture_supp/Compare_mv.pdf}
% % \vspace{-3mm}
%     \caption{
%     \textbf{Qualitative ablation study}. The videos are generated by taking textual descriptions
% and structural guidance as conditions.}
% %
% % Without the temporal coherence loss, the synthesized videos suffer from temporal inconsistency such as color shift.
%     %
%     % }
%     % \vspace{-2mm}
%     \label{fig:compare_mv}
%     }
% \end{figure*}
% % }]

\twocolumn[
% \maketitlesupplementary
{
% \maketitle
% \twocolumn[
        \centering
        \Large
        \textbf{\thetitle}\\
        \vspace{0.3em}Supplementary Material \\
        \vspace{1.5em}
        % {
        % \small
        % \iftoggle{cvprrebuttal}{}{
        %       Anonymous \confName~submission\\
        %       \vspace*{1pt}\\
        %       Paper ID \paperID \\
        %     }
        %     }
        % \vspace{1.5em}
       % ] %< twocolumn
% \maketitlesupplementary
    \centering
    \vspace{5pt}
    \includegraphics[width=1.00\textwidth]{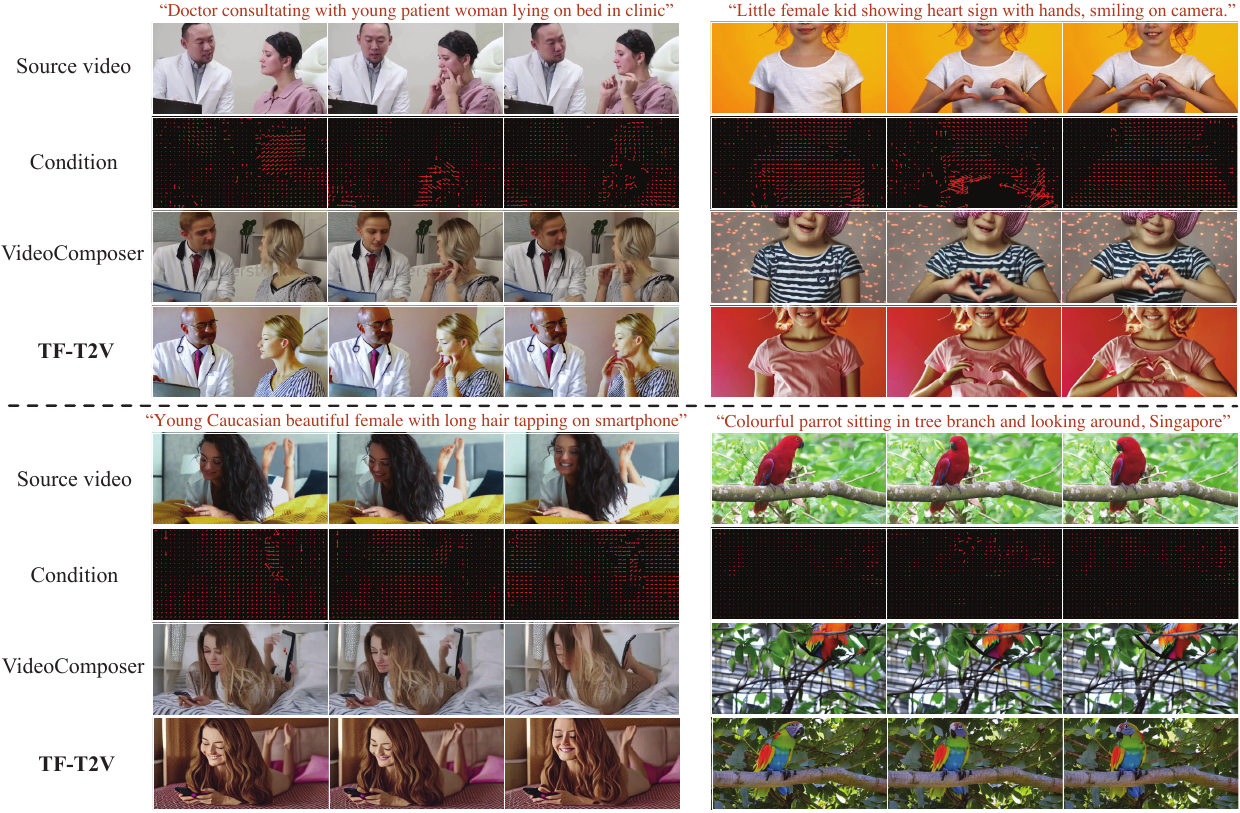}
    \vspace{-28pt}
    \captionof{figure}{
        \textbf{Qualitative comparison on compositional motion-to-video synthesis}. The videos are generated by taking textual descriptions
and motion vectors as conditions. Compared to VideoComposer, TF-T2V produces more realistic and appealing results.
        %
        % Since only language-free videos are required, High-definition videos can be easily generated.
        % High-definition videos can be 
        %
    }
    \label{first_figure_supp}
    \vspace{18pt}
    }
]

% \begin{figure*}[t]
% % \twocolumn[
% {
% % \maketitle
% % \maketitlesupplementary
%     \centering
% \includegraphics[width=1.00\textwidth]{Picture_supp/Compare_mv_small.pdf}
%     \vspace{-24pt}
%     \captionof{figure}{
%         \textbf{Qualitative comparison on compositional motion-to-video synthesis}. The videos are generated by taking textual descriptions
% and motion vectors as conditions. Compared to VideoComposer, TF-T2V produces more realistic and appealing results.
%         %
%         % Since only language-free videos are required, High-definition videos can be easily generated.
%         % High-definition videos can be 
%         %
%     }
%     \label{first_figure_supp}
%     }
% \end{figure*}

% Due to page limitations, we move some parts
Due to the page limit of the main text, we add more details and experimental results in this appendix.
%
% In addition, 
Besides,
limitations and future work will also be discussed.

\begin{table}[t]
    % \vspace{-1.1em}
    \caption{
        {
        Ablation study on different training manners. 
    }}
     \label{tab:compare_with_train_manner}
     \vspace{-3mm}
    \tablestyle{4pt}{0.95}
    % \vspace{-5pt}
    \centering
    \setlength{\tabcolsep}{10pt}{
   % \begin{tabular}{@{}lc@{}}
   % \resizebox{\textwidth}{20mm}{
   % \scalebox{0.98}{
      \begin{tabular}{l|ccc}
    \shline
        Setting       &  FID ($\downarrow$) &   FVD  ($\downarrow$)  & CLIPSIM ($\uparrow$) \\
        \shline

        Separately   &  9.22 & 503 & 0.2905 \\
        \rowcolor{Gray}
        Jointly  &   \textbf{8.19}  & \textbf{441}  & \textbf{0.2991}  \\
        % \shline
       \shline
    \end{tabular}
    }
    % \vspace{-10pt}
    \vspace{-3mm}
% \end{wraptable}
\end{table}

\section{More experimental details}

In \cref{first_figure_supp}, we show the comparison on compositional motion-to-video synthesis.
\method achieves more appealing results than the baseline VideoComposer.
Following prior works, 
we use an off-the-shelf pre-trained variational
autoencoder (VAE) model from Stable Diffusion 2.1 to encode the latent features.
The VAE encoder has a downsample factor of 8.
% The model structure of A is basically the same as that of B and C to facilitate fair comparison.
%
In the experiment,
the network structure of \method is basically consistent with the open source ModelScopeT2V and VideoComposer to facilitate fair comparison.
Note that \method is a plug-and-play framework that can also be applied to other text-to-video generation and controllable video synthesis methods.
%
% To verify the high-definition generation capability of \method, we train 
% To verify that \method can be extended to high-definition video generation, we also use text-free videos to train a high-resolution text-to-video model, such as $896\times512$. For high-definition compositional video synthesis, we synthesize $1280\times640$ and $1280\times768$ videos to demonstrate the excellent application potential of our method.
%
% \method support different resolutions as input
% As for human evaluation, we randomly generated 100 videos and asked users to 
% [Details about user study]
For human evaluation, we randomly generate 100 videos and ask users to rate and evaluate them. The highest score for each evaluation content is 100\%, the lowest score is 0\%, and the final statistical average is reported.

\section{Additional ablation study}
% separately
\noindent 
\textbf{Effect of joint training.}
% and eliminate the domain gap
In our default setting, we jointly train the spatial and temporal blocks in the video diffusion model to fully exploit the complementarity between image and video modalities.
An alternative strategy is to separate the spatial and temporal modeling into sequential two stages.
% temporal 
We conduct a comparative experiment on these two strategies in \cref{tab:compare_with_train_manner}.
The results demonstrate the rationality of joint optimization in \method.

\begin{figure}[t]
% \twocolumn[
{
% \maketitle
% \maketitlesupplementary
    \centering
\includegraphics[width=0.47\textwidth]{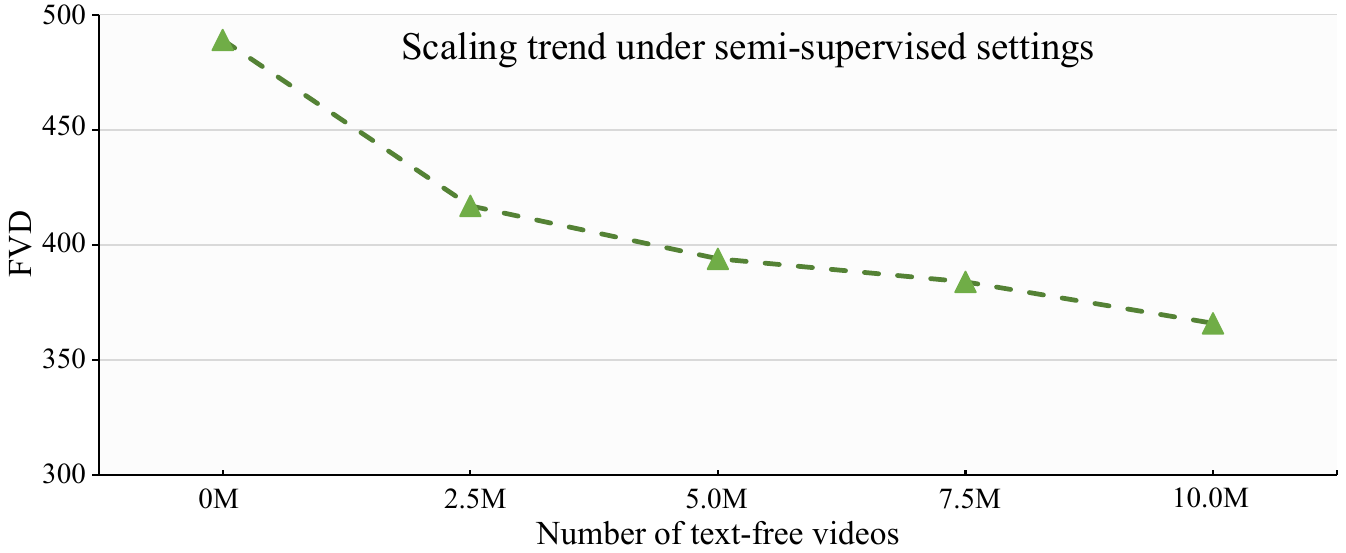}
\vspace{-2mm}
    \caption{
    \textbf{Scaling trend under semi-supervised settings}.  In the experiment, labeled
WebVid10M and text-free videos from Internal10M are leveraged.}
%
% Without the temporal coherence loss, the synthesized videos suffer from temporal inconsistency such as color shift.
    %
    % }
    % \vspace{-2mm}
    \label{fig:scaling_trend_semi}
    }
\end{figure}

\vspace{1mm}
\noindent 
\textbf{Scaling trend under semi-supervised settings.}
In \cref{fig:scaling_trend_semi}, we vary the number of text-free videos and explore the scaling trend of \method under the semi-supervised settings.
%
% [The impact of the amount of text-free data on performance]
%
From the results, we can observe that FVD ($\downarrow$) gradually decreases as the number of text-free videos increases, revealing the strong scaling potential of our \method.
%
% We hope that our \method can provide inspirations for future work to exploring text-free videos at large scales.
% We hope that our approach can provide inspiration for future work exploring the use of large-scale text-free videos to facilitate video synthesis.

% or use feature pooling

\begin{figure*}[h]
% \twocolumn[
{
% \maketitle
% \maketitlesupplementary
    \centering
\includegraphics[width=0.99\textwidth]{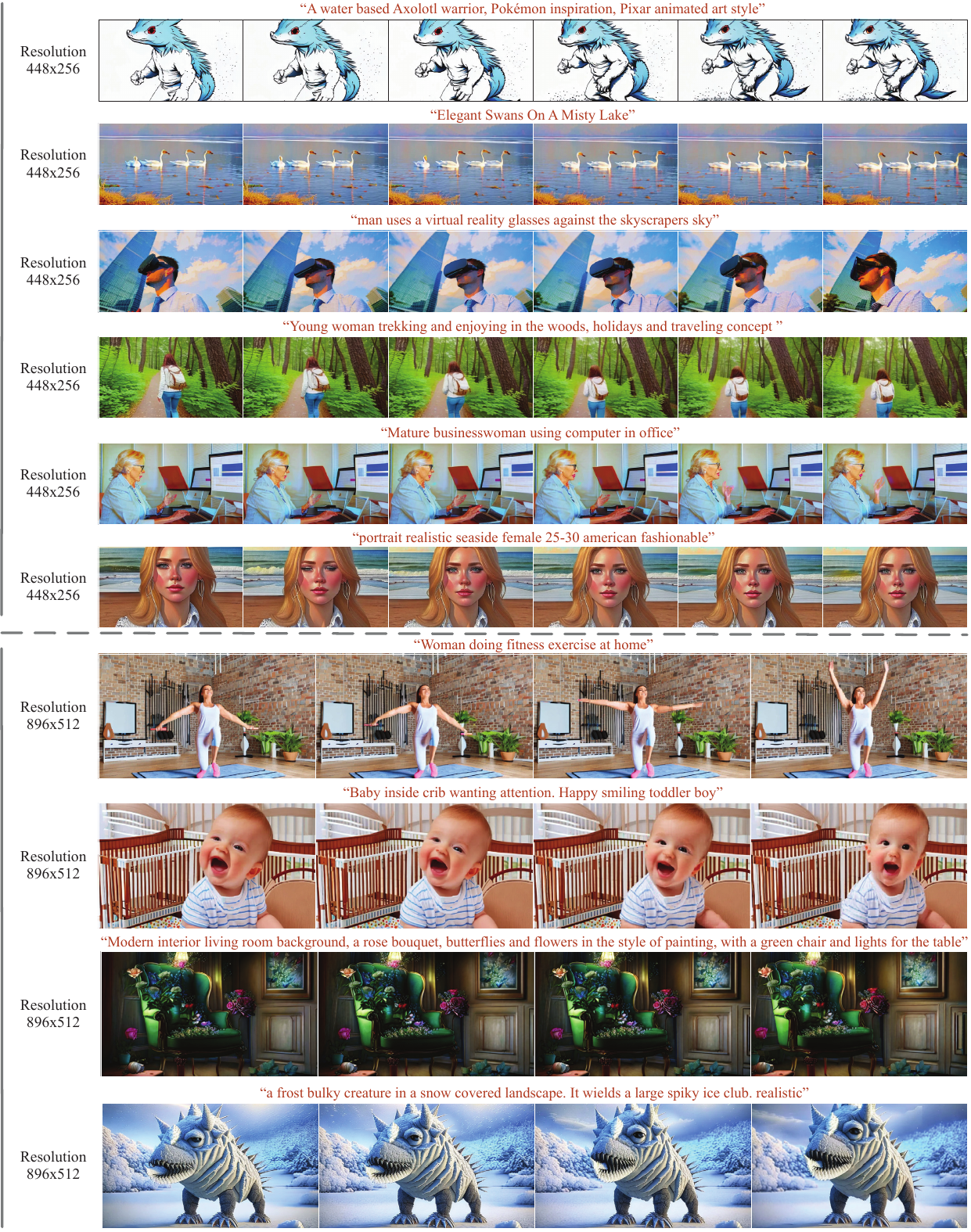}
% \vspace{-3mm}
    \caption{
    \textbf{More video results on text-to-video generation} without training on any video-text pairs. 
    % The videos are generated by taking textual descriptions and structural guidance as conditions.
    %
    Two resolution types of video are generated, $448\times256$ and $896\times512$ respectively.
}
%
% Without the temporal coherence loss, the synthesized videos suffer from temporal inconsistency such as color shift.
%
    %
    % }
    % \vspace{-2mm}
    \label{fig:compare_t2v_supp}
    }
\end{figure*}

\begin{figure*}[h]
% \twocolumn[
{
% \maketitle
% \maketitlesupplementary
    \centering
\includegraphics[width=0.99\textwidth]{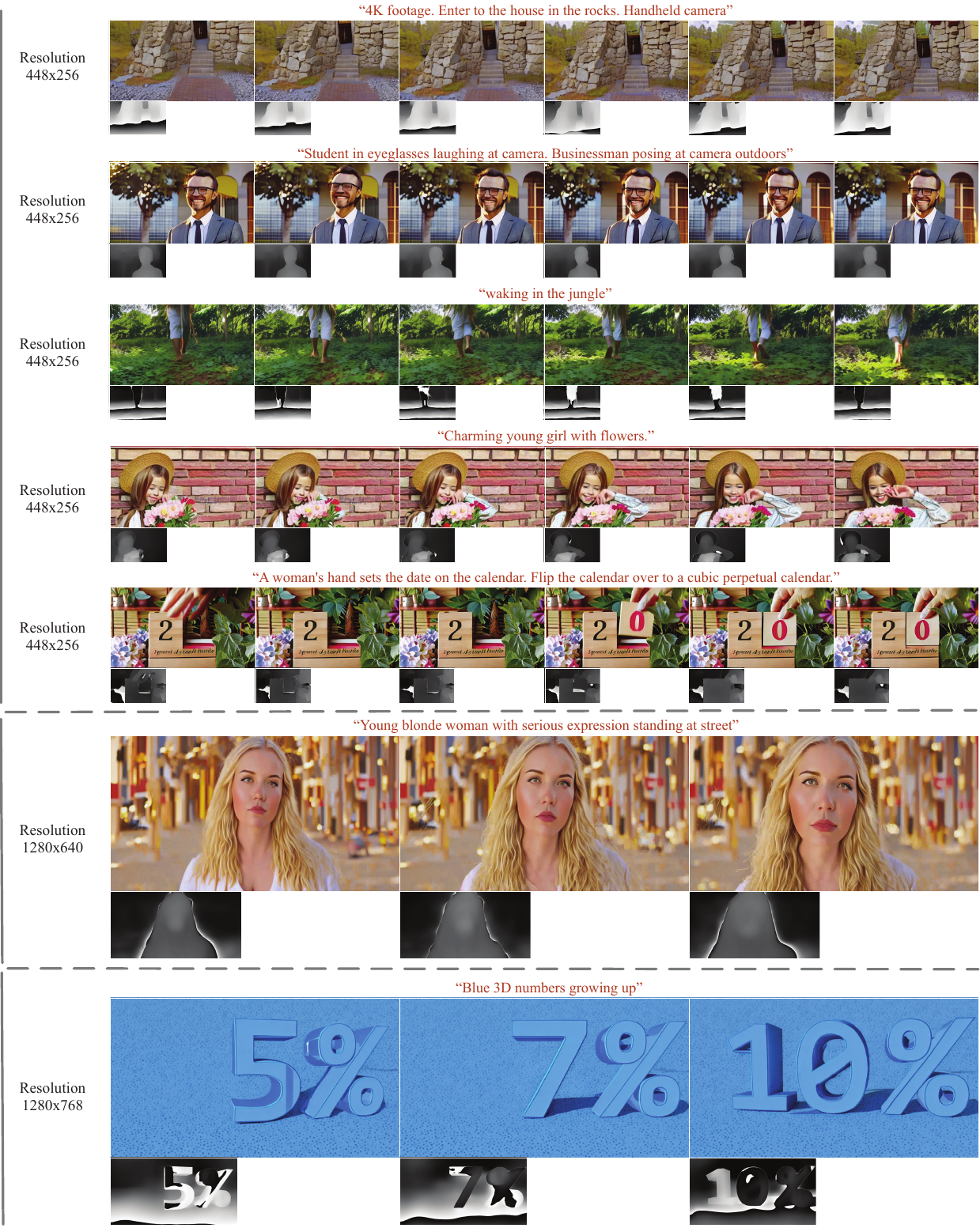}
% \vspace{-3mm}
    \caption{
    \textbf{More video results on compositional depth-to-video synthesis} without training on any video-text pairs. 
    %
%     The videos are generated by taking textual descriptions
% and structural guidance as conditions.
Three resolution types of video are generated, $448\times256$, $1280\times640$ and $1280\times768$ respectively.
}
%
% Without the temporal coherence loss, the synthesized videos suffer from temporal inconsistency such as color shift.
    %
    % }
    % \vspace{-2mm}
    \label{fig:compositional_supp_depth}
    }
\end{figure*}

\begin{figure*}[h]
% \twocolumn[
{
% \maketitle
% \maketitlesupplementary
    \centering
\includegraphics[width=0.99\textwidth]{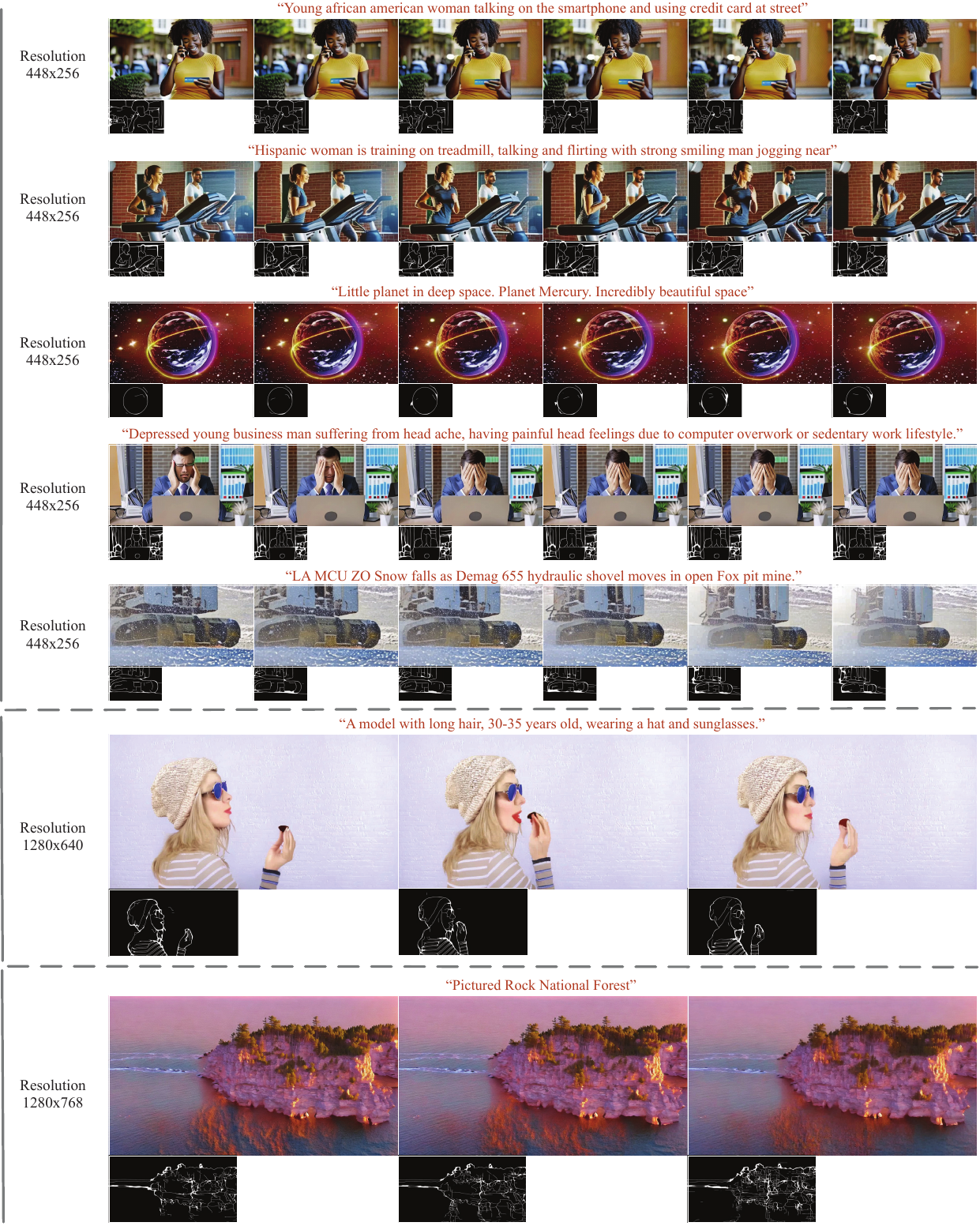}
% \vspace{-3mm}
    \caption{
    \textbf{More video results on compositional sketch-to-video synthesis} without training on any video-text pairs. 
%     The videos are generated by taking textual descriptions
% and structural guidance as conditions.
%
Three resolution types of video are generated, $448\times256$, $1280\times640$ and $1280\times768$ respectively.
}
%
% Without the temporal coherence loss, the synthesized videos suffer from temporal inconsistency such as color shift.
    %
    % }
    % \vspace{-2mm}
    \label{fig:compositional_supp_sketch}
    }
\end{figure*}

\section{More experimental results}

% In addition to generating $448\times256$ videos, our method can be easily applied to the field of high-definition video synthesis.
% further
To further verify that \method can be extended to high-definition video generation, we leverage text-free videos to train a high-resolution text-to-video model, such as $896\times512$. 
As shown in \cref{fig:compare_t2v_supp}, we can notice that in addition to generating $448\times256$ videos, our method can be easily applied to the field of high-definition video synthesis.
For high-definition compositional video synthesis, we  additionally synthesize $1280\times640$ and $1280\times768$ videos to demonstrate the excellent application potential of our method.
The results are displayed in \cref{fig:compositional_supp_depth} and \cref{fig:compositional_supp_sketch}.

\section{Limitations and future work}

% We hope our work can
% shed light on future research on

% scalability
% Due to computational constraints, we scaled the 
In this paper, we only doubled the training set to explore scaling trends due to computational resource constraints, leaving the scalability of larger scales (such as $10\times$ or $100\times$) unexplored. We hope that our approach can shed light on subsequent research to explore the scalability of harnessing text-free videos.
% will inspire subsequent approaches.
% The second limitation is the lack of exploration of long videos.
% The second limitation of this work is the ability to process long videos.
The second limitation of this work is the lack of exploration of processing long videos. 
In the experiment,
we follow mainstream techniques and sample 16 frames from each video clip to train our \method for fair comparisons. Investigating long video generation with text-free videos is a promising direction.
% Under the constraints of memory and computation time, \method models usually try to select video frames to reduce the computational burden.
%
% In addition, we find that if the input textual prompts contain  some fine attributes, such as ``pink pants", ``blue hair", etc., our \method may cause attribute shift problems and struggle to accurately synthesize the desired video.
%
% Investigating how accurately attribute information can be injected into a model through is a worthwhile research question
%
% Studying how to accurately inject attribute information into the video diffusion model is a worthwhile research subject.
In addition, we find that if the input textual prompts contain some temporal evolution descriptions, such as ``from right to left", ``rotation", etc., the text-free \method may fail and struggle to accurately synthesize the desired video. In the experiment, even though we noticed that semi-supervised \method helped alleviate this problem, it is still worth studying and valuable to precisely generate satisfactory videos with high standards that conform to the motion description in the given text.
% it is still challenging to precisely generate satisfactory videos that match the motion descriptions in the text to a high standard.

\end{document}